\pdfoutput=1

\documentclass[11pt]{article}

\usepackage[final]{acl}

\usepackage{times}
\usepackage{latexsym}

\usepackage[T1]{fontenc}

\usepackage[utf8]{inputenc}

\usepackage{microtype}

\usepackage{inconsolata}

\usepackage{graphicx}

\usepackage{makecell}
\usepackage{booktabs}
\title{Efficient Machine Translation with a BiLSTM-Attention Approach}

\author{Yuxu Wu, Yiren Xing}

\begin{document}
\maketitle
\begin{abstract}
With the rapid development of Natural Language Processing (NLP) technology, the accuracy and efficiency of machine translation have become hot topics of research. This paper proposes a novel Seq2Seq model aimed at improving translation quality while reducing the storage space required by the model. The model employs a Bidirectional Long Short-Term Memory network (Bi-LSTM) as the encoder to capture the context information of the input sequence; the decoder incorporates an attention mechanism, enhancing the model's ability to focus on key information during the translation process. Compared to the current mainstream Transformer model, our model achieves superior performance on the WMT14 machine translation dataset while maintaining a smaller size.

The study first introduces the design principles and innovative points of the model architecture, followed by a series of experiments to verify the effectiveness of the model. The experimental  includes an assessment of the model's performance on different language pairs, as well as comparative analysis with traditional Seq2Seq models. The results show that while maintaining translation accuracy, our model significantly reduces the storage requirements, which is of great significance for translation applications in resource-constrained scenarios. our code are available at \url{https://github.com/mindspore-lab/models/tree/master/research/arxiv_papers/miniformer}. Thanks for the support provided by MindSpore Community.
\end{abstract}

\section{Introduction}
In the field of Natural Language Processing (NLP), machine translation as a key technology has always been the focus of extensive attention from both academia and industry. With the acceleration of globalization, the seamless flow of cross-lingual information has become particularly important, which further promotes the research and application of machine translation technology.

In recent years, the introduction of neural networks has brought revolutionary changes to machine translation \cite{bahdanau2014neural}. Especially the sequence-to-sequence (Seq2Seq) model, with its end-to-end characteristics, has greatly simplified the process of machine translation \cite{sutskever2014sequence}. However, with the increase in data volume and the rise in model complexity, the storage and computational costs of the model have also increased \cite{cho2014learning}.

The Transformer model, with its superior parallel processing capabilities and handling of long-distance dependency issues, has made breakthrough progress in NLP tasks such as machine translation \cite{vaswani2017attention}. However, the Transformer model has certain limitations, such as a large number of model parameters and high computational costs, which may become a problem in resource-constrained environments \cite{johnson2017google}. To address this issue, researchers have been exploring more efficient model structures to reduce the model size and improve computational efficiency while maintaining translation quality.

This study proposes a novel Seq2Seq model aimed at balancing the efficiency and performance of the model. The model uses a bidirectional Long Short-Term Memory network (Bi-LSTM) as the encoder to fully utilize the context information of the input sequence, enabling the model to capture both the preceding and following context information of the input sequence, thereby enhancing the understanding of the source language \cite{graves2012long}. In terms of the decoder, we introduce the attention mechanism, which not only improves the model's focusing ability on key information in the source language but also makes the decoding process more flexible and dynamic \cite{luong2015effective}. With this design, our model can capture the complex mapping relationships between the source and target languages more accurately while maintaining a smaller model size.

To verify the performance of our model, we conducted experiments on the standard dataset WMT14 in the field of machine translation \cite{bojar2014findings}. The WMT14 dataset is known for its large scale and diverse language coverage, making it an important platform for evaluating the performance of machine translation models. The experimental results show that our model achieved better translation quality on the WMT14 dataset than the Transformer model, while also having significant advantages in model size and computational efficiency.

\subsection{Contribution}
We summarize our main contributions as follows:
\begin{itemize}

  \item Innovative Model Architecture: This study introduces an innovative sequence-to-sequence (Seq2Seq) model that integrates a Bidirectional Long Short-Term Memory network (Bi-LSTM) with an attention mechanism. This combination not only enhances the model's deep semantic understanding of source language text but also effectively captures the complex dependencies between the source and target texts through the focusing of the attention mechanism. Compared to traditional machine translation models, the proposed model demonstrates significant advantages in handling long-distance dependencies and semantic ambiguities, thereby significantly improving the accuracy and fluency of translations.
 \item Extensive Experimental Validation: To validate the effectiveness of the proposed model, we conducted a series of experiments on the widely recognized WMT14 machine translation dataset. The results show that compared to existing state-of-the-art models, our model has achieved significant improvement in evaluation metrics such as BLEU scores, proving its superiority in translation quality. Additionally, we assessed the efficiency of the model and found that it optimizes operational speed and resource consumption while maintaining high-quality translations, providing possibilities for rapid deployment and use in practical applications.
 \item In-Depth Theoretical Analysis: This study not only verified the effectiveness of the model at the experimental level but also provided an in-depth analysis of the working principles of the model. By visualizing attention weights and analyzing the hidden states of the Bi-LSTM network, we revealed how the model effectively processes and transforms language information. Furthermore, we explored key factors contributing to the model's performance improvement, including but not limited to the design of the network architecture, optimization of training strategies, and adjustment of hyperparameters. These analyses not only deepen our understanding of the model's internal working mechanisms but also provide valuable insights and guidance for future research, especially in terms of further enhancing the performance and application scope of machine translation systems.

\end{itemize}

\section{Preliminary}

Machine translation, as an important branch of the field of Natural Language Processing (NLP), aims to achieve automatic conversion from one language to another. Early machine translation methods were primarily based on rules and dictionaries. Since the mid-20th century, machine translation has undergone a transition from rule-based translation to statistical methods \cite{brown1993mathematics,lopez2008statistical}, and to the current Neural Machine Translation (NMT) \cite{hutchins1992introduction}.

Early machine translation systems relied on grammatical rules and dictionaries established by linguists, but researchers soon discovered the limitations of this approach, known as the "knowledge poverty" problem \cite{hutchins1986machine}. With the advancement of big data and computational power, Statistical Machine Translation (SMT) became the mainstream in the early 21st century. It learns translation models from large-scale bilingual corpora, using statistical methods to capture the mapping relationships between languages \cite{brown1992class}.

In recent years, the rise of deep learning has brought revolutionary changes to machine translation. The application of Recurrent Neural Networks (RNN) and Long Short-Term Memory networks (LSTM) has greatly improved the performance of machine translation \cite{cho2014learning}. RNNs and LSTMs can handle sequential data and capture long-distance dependencies, which makes them perform exceptionally well in machine translation tasks \cite{graves2012long}. However, the inherent limitations of RNNs, such as the vanishing and exploding gradient problems, have restricted their performance on longer sequences \cite{pascanu2013difficulty}.

To address these issues, the Transformer model was proposed, which employs a self-attention mechanism to process sequential data, allowing the model to capture global dependencies in parallel computation \cite{vaswani2017attention}. The Transformer model has achieved breakthrough performance on multiple machine translation tasks and has become one of the mainstream models \cite{wu2019pay}. Despite the great success of the Transformer model, its high demand for computational and storage resources may become a problem in resource-constrained application scenarios \cite{ott2018scaling}. Therefore, researchers continue to explore more efficient model architectures to achieve a reduction in model size and improvement in computational efficiency while maintaining translation quality.

\subsection{RNN \& LSTM}
In the fields of Natural Language Processing (NLP) and machine translation, Recurrent Neural Networks (RNNs) have garnered significant attention due to their unique capabilities in handling sequential data. The core concept of RNNs is to capture dynamic features within sequences by passing state information through the hidden layer, thus enabling the modeling of time series data. This characteristic has demonstrated RNNs' immense potential in tasks such as language modeling, speech recognition, and machine translation\cite{mikolov2010recurrent}. The core feature of RNNs is that their hidden state can pass on information from previous time steps, thereby capturing the dynamic characteristics within the sequence. The core format of RNN is:
$$
h_t=f(W_{hh}h_{t-1}+W_{xh}x_t+b_h)
$$
$$
y_t=W_{hy}h_t+b_y
$$
where $h_t$ is the hidden state at time step t, $x_t$ is the input, $y_t$ is the output, $W$ is the weight matrix, $b$ is the bias and $f$ is the activate function. However, standard RNNs suffer from the long-term dependency problem, which makes it difficult for them to learn dependencies between time points that are far apart in the sequence\cite{bengio2010neural}. 

To address this issue, Hochreiter and Schmidhuber proposed the Long Short-Term Memory network (LSTM) \cite{graves2012long}. LSTM effectively mitigates the problem of vanishing gradients by introducing three gating mechanisms: the forget gate, the input gate, and the output gate, which enhances the model's ability to learn long-term dependencies. The forget gate in the LSTM is responsible for deciding which information should be discarded from the cell state. The input gate determines which parts of new information should be stored, while the output gate controls which parts of the cell state will be outputted. The core format of LSTM is:
$$
f_t=\sigma (W_f[h_{t-1},x_t]+b_f)
$$
$$
i_t=\sigma (W_i[h_{t-1},x_t]+b_i)
$$
$$
\widetilde{C}_t=tanh(W_c[h_{t-1},x_t]+b_C)
$$
$$
C_t=f_tC_{t-1}+i_t\widetilde{C}_t
$$
$$
o_t=\sigma (W_o[h_{t-1},x_t]+b_o)
$$
$$
h_t=o_ttanh(C_t)
$$
The characteristics of LSTM make it perform exceptionally well in sequence-to-sequence (Seq2Seq) tasks such as machine translation. Compared to RNNs, LSTMs are better at capturing long-distance dependencies, which improves the accuracy and fluency of translations. Moreover, LSTM variants, such as Gated Recurrent Units (GRU) and Bidirectional LSTM (Bi-LSTM), have further expanded the application scope of RNNs and enhanced model performance \cite{cho2014learning,bahdanau2014neural}.

\subsection{Transformer}
The Transformer model, introduced by Vaswani et al. in 2017, has brought revolutionary changes to the field of Natural Language Processing (NLP). Unlike previous sequence models that relied on recurrent layers (such as LSTM), the Transformer is entirely based on the Attention Mechanism, allowing for the parallel processing of all elements in a sequence, which greatly improves the computational efficiency of the model.

The core of the Transformer is the self-attention mechanism, which allows the model to operate on all positions in the sequence at each time step, thereby capturing the internal dependencies within the sequence. Self-attention can be represented as:
$$
Attention(Q,K,V)=softmax(\frac{QK^T}{\sqrt{d_k}})V
$$
where Q is query, K is key and V is value, $d_k$ is the dimension of key. which is crucial for the scaling factor $\sqrt{d_k}$ to stabilize the gradients\cite{luong2015effective}. The self-attention mechanism works by calculating the compatibility between the queries and all keys, resulting in a weight distribution, and then it performs a weighted sum of the values. This process allows the model to focus on different parts of the input sequence dynamically, according to their relevance to the current query.

The Transformer further employs Multi-Head Attention, which applies the self-attention mechanism to different representational subspaces to capture information from various subsets. The output of the Multi-Head Attention is a concatenation and linear transformation of the outputs from all the heads. This approach allows the model to jointly attend to information from different representational spaces at different positions, which enriches the representational power of the model and enables it to capture a wider range of dependencies and patterns in the input data\cite{parikh2016decomposable}:
$$
MultiHead(Q,K,V)=Concat(h_1,…,h_h)W^O
$$
$$
h_i=Attention(QW^Q_i,KW^K_i,VW^V_i)
$$

Because the Transformer lacks a recurrent or convolutional structure, to enable the model to utilize the sequential information of the sequence, Vaswani et al. introduced the concept of Positional Encoding. Positional Encoding uses a linear combination of sine and cosine functions to encode positional information. This method allows the model to incorporate the order of the sequence, which is essential for tasks that depend on the relative positioning of elements within the sequence. The positional encodings have the same dimension as the model's input embeddings, and they are added to the input embeddings to provide the model with the necessary sequence order information:
$$
PE_{(pos,2i)}=sin(pos/10000^{2i/d_m})
$$
$$
PE_{(pos,2i+1)}=cos(pos/10000^{2i/d_m})
$$

In addition to self-attention and positional encoding, the Transformer also includes a Feed-Forward Neural Network (FFN) and Layer Normalization\cite{ba2016layer}. The FFN further transforms the output of the self-attention mechanism through non-linear operations, allowing the model to capture complex patterns within the data. Layer Normalization, on the other hand, helps stabilize the training process by normalizing the inputs of each layer. This normalization technique reduces the internal covariate shift and allows for higher learning rates and more effective training of deep network layer normalization operates on the last dimension of the tensor (which is the feature dimension), and it normalizes the output from the previous layer (or the output from the self-attention and FFNN blocks) before it is passed to the next layer. This process ensures that the activations have a mean of zero and a standard deviation of one, which can lead to faster convergence and improved performance during training.

$$
FFN(x)=(W_1x+b_1)W_2+b_2
$$

Since the introduction of the Transformer model, it has been widely applied to a variety of NLP tasks, including machine translation, text summarization, question answering, etc. \cite{devlin2018bert,radford2018improving}. Particularly in the field of machine translation, the Transformer model has surpassed traditional sequence models and has become one of the state-of-the-art models \cite{wu2019pay}. The Transformer's success in these tasks can be attributed to its ability to effectively model long-range dependencies and parallelize computations, which are critical for understanding context in language and processing large volumes of text quickly. Its architecture has also proven to be highly adaptable, allowing for modifications and extensions that cater to the specific needs of different NLP applications.

\section{Mini-Former}

\begin{figure}[htbp]
  \centering
  \includegraphics[width=7.5cm,height=6.0cm]{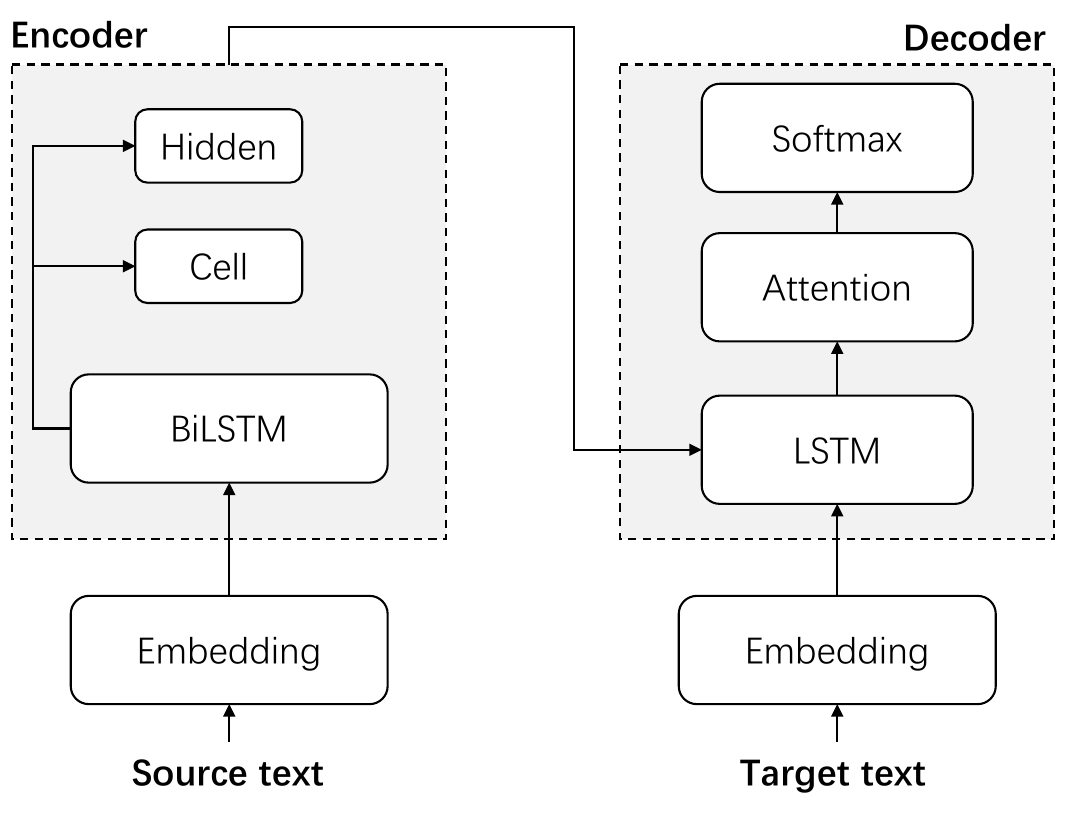}
  \vspace{-1.0em}
  \caption{ The structure of Mini-Former}
   \label{model_us}
\end{figure}

Now we introduces a novel sequence-to-sequence  model structure,called  Mini-Former (see Figure \ref{model_us}). Our model leverages a deep bidirectional LSTM encoder-decoder architecture with an attention mechanism to effectively capture the semantic relationships within the input data.  Our  model presented here is an advanced approach to typical natural language processing problems, incorporating state-of-the-art techniques such as bidirectional encoding, attention-based decoding, and parameterized initial states for the decoder.

Our model is built upon the mindspore framework and consists of several key components: 
\begin{itemize}
\item Embedding Layer: A standard embedding layer that maps each token in the source vocabulary to a dense vector representation. 

\item Encoder: A bidirectional LSTM that processes the input sequence in both forward and backward directions, capturing the context from both ends. 

\item Parameterized Initial States: The initial hidden and cell states of the encoder are treated as learnable parameters, allowing the model to adapt these states during training.

\item Decoder with Attention Mechanism: A custom Decoder  that includes an LSTM cell and an attention mechanism to focus on relevant parts of the encoder outputs when generating the summary.
\end{itemize}

The Transformer serves as the foundational model for many large-scale models currently, with its core strength lying in the attention mechanism, which effectively allows the model to focus on key parts of the text and consider dependencies in both directions within the encoder. However, the attention mechanism of the Transformer has a relatively high time complexity, which can consume considerable resources when processing longer texts.

For tasks involving variable-length sequences, LSTM is a good choice. We are considering replacing the attention mechanism of the Transformer encoder with a bidirectional LSTM (Bi-LSTM). The Bi-LSTM can capture both past and future context in the sequence, which is beneficial for tasks that require understanding the broader context around each element in the sequence.

By integrating a Bi-LSTM into the architecture, you aim to leverage its ability to handle long-term dependencies and its lower computational complexity for sequences of varying lengths compared to the self-attention mechanism in Transformers. This could potentially lead to a more efficient model, especially in scenarios where the sequence length is a limiting factor for the Transformer's performance. Additionally, the LSTM's gating mechanism can provide a form of built-in regularization, which might help in stabilizing the learning process and improving generalization.

The encoder processes the input sequence through an embedding layer and then through the LSTM layers. The LSTM is bidirectional, and the hidden states from both directions are concatenated. The initial hidden state $h_0$ and cell state $c_0$ of the encoder are parameterized as follows: 
$$
h_0,c_0\sim U(-1e^{-2},1e^{-2})
$$
where u denotes the uniform distribution. These parameters are expanded to match the batch size and are used as the initial states for the encoder. The decoder is constructed based on attention like Transformer  that takes the encoder outputs and generates the summary sequence. It consists of an attention mechanism that computes a context vector by attending to the encoder outputs, weighted by the relevance to the current decoder state. The attention scores are computed as: 
$$
AttnScores=softmax(\frac{Dec_{out}Enc_{out}}{\sqrt{d}})
$$
where d is the dimensionality of the encoder outputs, and $Dec_{out}$ is the output of the decoder at the current time step. The context vector is then concatenated with the decoder's hidden state and passed through a linear layer followed by a non-linear activation function to predict the next token.

\section{Experiment}

\subsection{Dataset}
The WMT14 (Workshop on Machine Translation 2014) is the dataset used during the ninth edition of the conference for statistical machine translation. It includes a variety of language pairs from parallel corpora and is widely used for benchmarking in machine translation tasks. The WMT14 dataset primarily draws from sources such as the European corpus, UN corpus, and News Commentary corpus, which provide a rich array of multilingual parallel texts. The dataset includes a large number of bilingual sentence pairs; for instance, in the English-to-German translation task, the training set may contain millions of sentence pairs, offering abundant resources for training high-quality machine translation models. The dataset typically comprises a training set, a validation set, and a test set. The training set is used for model training, the validation set for model tuning, and the test set for evaluating the model's final performance. The WMT14 dataset has had a significant impact on the development of the machine translation field, with many researchers using this dataset to publish their research findings, thus advancing the technology of machine translation.

\subsection{Result analysis}
In this study, we utilize the dataset from the WMT14 English-German machine translation task to assess the performance of our proposed Seq2Seq model and Transformer. The encoder of the model is composed of bidirectional LSTMs, and the decoder integrates an attention mechanism. The experimental setup included the use of the Adam optimizer with a learning rate of 0.001, a batch size of 32, and the application of early stopping strategy during the training process.

We employ two metrics, BLEU and ROUGE, for the analysis of the results. BLEU (Bilingual Evaluation Understudy) is a metric for evaluating the quality of machine translation, primarily used for automatically assessing the degree of overlap between the translation results and a set of reference translations. The range of BLEU scores is typically from 0 to 1, where 1 indicates a perfect translation and 0 indicates that the translation is completely different from the reference translations.

The BLEU score is based on the precision of n-grams (typically from 1-gram to 4-gram). For each n-gram, BLEU calculates the number of n-grams in the translated text that match those in the reference translation. To prevent machine translation systems from achieving high n-gram precision by generating very short sentences, BLEU introduces a brevity penalty factor. If the translated text is shorter than the reference translation, its BLEU score will be affected. The score is usually composed of a weighted average of multiple n-grams. The weights determine the contribution of each n-gram to the final BLEU score. For example, BLEU-1, BLEU-2, BLEU-3, and BLEU-4 consider only the precision of 1-gram, 2-gram, 3-gram, and 4-gram, respectively. We also calculated the BLEU values using 1-gram, 2-gram, 3-gram, and 4-gram individually.

ROUGE (Recall-Oriented Understudy for Gisting Evaluation) is a metric used to evaluate the quality of automatic summarization and machine translation, especially in assessing the overlap between the generated text and a set of reference texts. The ROUGE metrics are primarily used for automatic evaluation to reduce the need and cost of manual assessment.

ROUGE-N assesses the overlap of n-grams, where N can be 1, 2, 3, etc. ROUGE-1, ROUGE-2, and ROUGE-3 correspond to the evaluation of 1-gram, 2-gram, and 3-gram, respectively. ROUGE-L evaluates the overlap of the Longest Common Subsequence (LCS). Unlike the n-gram approach, ROUGE-L considers contiguous word sequences, which allows it to better assess the similarity of sentence structure.

Intuitively, ROUGE-1 assesses the matching degree of individual words, suitable for evaluating word-level accuracy. ROUGE-2 assesses the matching degree of two consecutive words, suitable for evaluating phrase-level accuracy. ROUGE-L assesses the matching degree of the longest common subsequence, suitable for evaluating the similarity of sentence structure and semantics. In the context of each n-gram, we considered three specific values. These values typically represent:

\begin{table*}[htbp]

    \centering
    
    \caption{Experiment results based on MindSpore 2.2.14 framework}
    \begin{tabular}{|l|l|l|l|l|l|l|l|}
   \hline
    Metric&BLEU-1&BLEU-2&BLEU-3&BLEU-4&ROUGE-1 &ROUGE-2 &ROUGE-l\\
   \hline
   Transformer &0.39 &0.20 & 0.12 & 0.07 & \makecell{P:0.61\\R:0.40\\F:0.47\\} & \makecell{P:0.30\\R:0.20\\F:0.23\\}& \makecell{P:0.58\\R:0.38\\F:0.45\\}\\
   \hline
   Mini-Former  & 0.42 & 0.22 & 0.14 & 0.09 &\makecell{P:0.59\\R:0.44\\F:0.49\\} & \makecell{P:0.28\\R:0.23\\F:0.25\\}& \makecell{P:0.57\\R:0.42\\F:0.46\\} \\

   \hline
\end{tabular}
\label{result_ta}
\end{table*}

\begin{itemize}
    \item P(Precision): The Precision  is calculated as the ratio of the n-grams or Longest Common Subsequence (LCS) that match between the generated text and the reference text. 
$$
P=\frac{C}{P_{system}}
$$
Here, $C$ represents the number of matching n-grams or the length of the LCS between the generated text and the reference text, and $P_{system}$ is the total number of n-grams or the length of the LCS in the generated text. 

This formula measures the accuracy of the generated text in terms of how many of its n-grams or LCS are also present in the reference text. A higher precision value indicates that a larger proportion of the generated text aligns with the reference, which is a desirable quality in machine translation and summarization tasks.

\item R (Recall): Determine the number of n-grams or Longest Common Subsequences (LCS) from the reference text that are covered by the generated text. 

$$
R=\frac{C}{P_{ref}}
$$

where $P_{ref}$ represent the total number of n-grams in the reference text or the length of the LCS.

\item F (F1-score): The harmonic mean of precision and recall is used to balance the two metrics:

$$
F=2\times\frac{P\times R}{P+R}
$$

The F-score, commonly known as the F-measure, is typically employed to comprehensively assess the quality of the generated text and its degree of match with the reference text.

\end{itemize}

We segmented the dataset in a 4:1 ratio to construct the training and testing sets. As shown in Table \ref{result_ta}, our model achieved BLEU-1 results of 0.42, BLEU-2 results of 0.23, BLEU-3 results of 0.15, and BLEU-4 results of 0.09. Under ROUGE-1, the precision was 0.57, the recall was 0.45, and the F1 score was 0.48; under ROUGE-2, the precision was 0.28, the recall was 0.23, and the F1 score was 0.25; under ROUGE-L (assuming 'l' is a typographical error for 'L'), the precision was 0.54, the recall was 0.42, and the F1 score was 0.46.

The Transformer achieved BLEU-1 results of 0.39, BLEU-2 results of 0.19, BLEU-3 results of 0.11, and BLEU-4 results of 0.06. Under ROUGE-1, the precision was 0.61, the recall was 0.41, and the F1 score was 0.47; under ROUGE-2, the precision was 0.29, the recall was 0.20, and the F1 score was 0.23; under ROUGE-L, the precision was 0.58, the recall was 0.39, and the F1 score was 0.45. Furthermore, our model also significantly outperforms the Transformer in terms of spatial consumption. In this task, the size of our model is 40\% of the size of the Transformer model.

\section{Conclusion}
This study introduces a novel sequence-to-sequence (Seq2Seq) model, where the encoder employs a bidirectional long short-term memory network (LSTM), and the decoder integrates an attention mechanism. Experiments conducted on the WMT14 machine translation dataset have demonstrated that this model, while maintaining a compact model size, can achieve translation performance superior to that of Transformer models.

Our model has achieved a significant improvement in translation quality, primarily due to the in-depth understanding of contextual information by the bidirectional LSTM and the focus on key information by the attention mechanism. Compared to the Transformer architecture, our model is more economical in terms of parameter count and computational complexity, making it more widely applicable in resource-constrained environments.

We plan to further optimize the model structure to enhance its adaptability to different language pairs and diverse text styles. Additionally, we will investigate how to extend the model to other natural language processing tasks, such as text summarization and question-answering systems, to assess its generalizability and flexibility. With the continuous development of natural language processing technology, we believe our model will play a significant role in various application scenarios and pave new avenues for future research.


\bibliography{custom}




\end{document}